# A NEW FACE DATABASE SIMULTANEOUSLY ACQUIRED IN VISIBLE, NEAR INFRARED AND THERMAL SPECTRUM

Virginia Espinosa-Duró (1), Marcos Faundez-Zanuy (1), Jiří Mekyska (2)
(1) EUP Mataró Tecnocampus, Avda. Ernest Lluch 32, 08302 MATARO (BARCELONA), SPAIN
(2) SPLab, Department of Telecommunications, Faculty of Electrical Engineering and Communication, Brno University of Technology
espinosa@eupmt.es, faundez@eupmt.es, j.mekyska@phd.feec.vutbr.cz

*Abstract* –

In this paper we present a new database acquired with three different sensors (visible, near infrared and thermal) under different illumination conditions. This database consists of 41 people acquired in four different acquisition sessions, five images per session and three different illumination conditions. The total amount of pictures is 7.380 pictures.

Experimental results are obtained through single sensor experiments as well as the combination of two and three sensors under different illumination conditions (natural, infrared and artificial illumination). We have found that the three spectral bands studied contribute in a nearly equal proportion to a combined system. Experimental results show a significant improvement combining the three spectrums, even when using a simple classifier and feature extractor. In six of the nine scenarios studied we obtained identification rates higher or equal to 98%, when using a trained combination rule, and two cases of nine when using a fixed rule.

*Keywords:* thermal image, visible image, near infrared image, face recognition data fusion.

## 1. INTRODUCTION

Face is one of the most challenging traits for biometric recognition [1]. Real-world tests of automated face recognition systems have not yielded encouraging results. For instance, face recognition software at the Palm Beach International Airport, when tested on fifteen volunteers and a database of 250 pictures, had a success rate of less than fifty percent and nearly fifty false alarms per five thousand passengers, which means two to three false alarms per hour per checkpoint [2]. Even when the face recognition task is performed by a human operator it is far from being perfect and errors exist. Humans can recognize faces from different views. However, they accurately do it only when the faces are well known to them.

The limits of human performance do not necessarily define upper bounds on what is achievable. Specialized identification systems, such as those based on novel sensors, may exceed human performance in particular settings [2]. For this reason, it is interesting to perform automatic experiments with images acquired with different sensors. To this aim we studied in our previous work [3-4] if there is complementary information when capturing an image with sensors that acquire the face in different frequency ranges.

A long standing focus of research in human perception and memory centers on the importance of the "average" or "prototype" in guiding recognition and categorization of visual stimuli. The theory is that categories of objects, including faces, are organized around a prototype or average. The idea is that the closer an item is to the category prototype, the easier it is to be recognized as an exemplar of the category. However, in biometric applications, the goal is not to detect a face in an image. We must determine if the face is known to us and whose face it is [2]. The authors of [5] found that faces rated as "typical" were recognized less accurately than faces rated as "unusual". In general, artists draw caricatures emphasizing facial features that are "unusual" in average population and are present in the caricaturized person. Despite the fact that caricatures are grotesque distortions of a face, they are often recognized more accurately and efficiently than actual images of the faces [5]. Computer generated caricatures likewise operate by comparing a face to the "average face" and then by exaggerating facial dimensions that deviate from the average [6]. The enhanced recognizability of caricatures by comparison to veridical faces may be due to the fact that the exaggeration of unusual features in these faces makes the person less confusable with other faces, and somehow or other "more like themselves" [2].

We would expect that the recognizability of individual faces should be predicted by the density of faces in the neighboring "face space". We might also expect that the face space should be most dense in the center near the average. The space should become progressively less dense as we move away from the average. If a computationally-based face space approximates the similarity space humans employ for face processing, we might expect that "typical" faces would be near the center of the space and that unusual or distinctive faces be far from the center. It follows, therefore, that computational models of face recognition will not perform equally well for all faces. These systems should, like humans, make more errors on typical faces than on unusual faces.

On the other hand, computer based systems can go beyond human limitations because they can "see" beyond cognitive limits. For this purpose, we created a new database. Although several databases exist that simultaneously acquire visible and near infrared [7] or visible and thermal images [8], we are not aware of an existing database containing visible, near infrared and thermal information simultaneously.

In this paper, we present a new database and we extract quantitative measurements in three spectral bands: visible, near infrared and thermal.

The paper is organized as follows: section two describes the new database. Section three presents the face segmentation and normalization procedure, feature extraction, classifier and experimental results. Section four summarizes the main conclusions.

## 2. FACE DATABASE

We have acquired a database of 41 people using three sensors simultaneously. Next sections describe the details of this new database acquired in visible, near infrared and thermal spectrum.

### 2.1 Acquisition scenario

Visible and thermal images have been acquired using a thermographic camera TESTO 880-3, equipped with an uncooled detector with a spectral sensitivity range from 8 to 14 μm and provided with a germanium optical lens, and an approximate cost of 8.000 EUR. For the NIR a customized Logitech Quickcam messenger E2500 has been used, provided with a Silicon based CMOS image sensor with a sensibility to the overall visible spectrum and the half part of the NIR (until 1.000 nm approximately) with a cost of approx. 30 EUR. We have replaced the default optical filter of this camera by a couple of Kodak daylight filters for IR interspersed between optical and sensor. They both have similar spectrum responses as showed in Figure 1 and are coded as wratten filter 87 and 87C, respectively. In addition, we have used a special purpose printed circuit board (PCB) with a set of 16 infrared leds (IRED) with a range of emission from 820 to 1.000 nm in order to provide the required illumination.

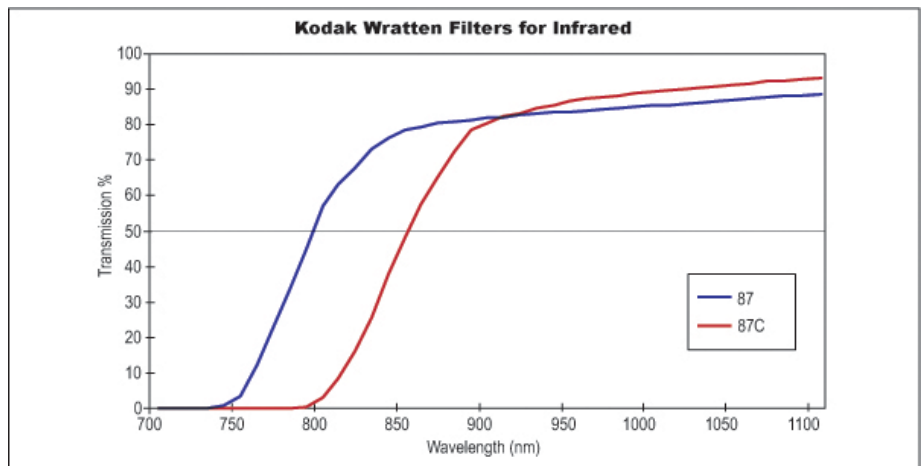

Figure 1. Spectral sensibility of the two visibly opaque infrared filters, specifically matched to our application.

The thermographic camera provides a resolution of 160x120 pixels for thermal images and 640x480 for visible images, whereas the webcam provides a still picture maximum resolution of 640x480 for near-infrared images and this has been the final resolution selected for our experiments.

The acquisition scenario is shown in figure 2. Figures 3a and 3b show a user in front of the acquisition system.

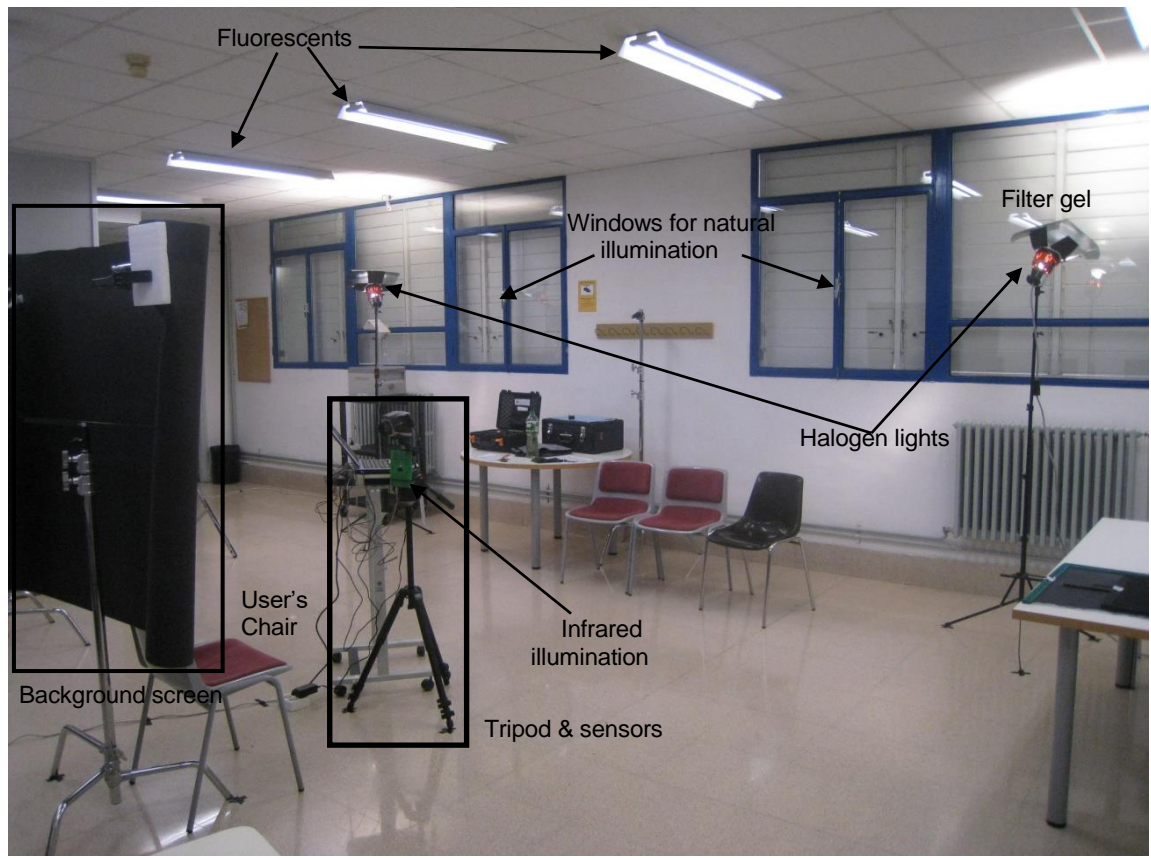

Figure 2. Overall multispectral face acquisition scenario.

A couple of halogen focus are placed 30°away from the frontal direction and about 3m away from the user match the artificial light of the room. Note that all the tripods and structures have fixed markings on the ground.

Additionally, the distance between the face of the user and the tripod that holds the sensors is 135cm. This is in order to minimize the inherent parallax error in short distances between visible and thermal images of thermal imager and also to obtain a similar field of view between these images and the near infrared images acquired with the customized webcam.

We have designed a background screen using a special stand kit which supports a roll of matt black paper. It is important to point out that this matt black background is mandatory behind the user in order to avoid undesirable thermal reflections from the operator, due to its well-known extra low albedo. This smooth background also facilitates the segmentation of the visible and NIR images.

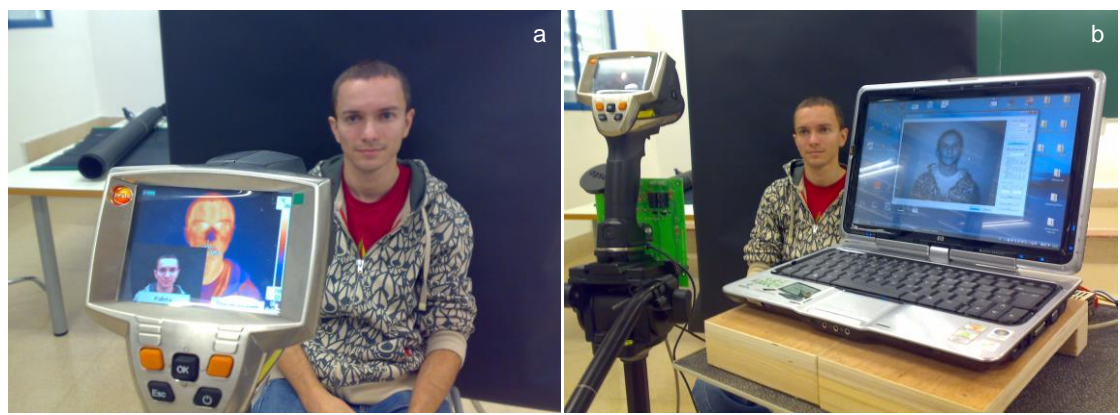

Figure 3. Acquisition scenario: the thermal camera (on the left) simultaneously acquires thermal and visible images in dual mode, while the printed circuit board with infrared leds and webcam must be connected to a laptop (on the right) to acquire the NIR image.

2.2 Lighting conditions.

In each recording session the images have been acquired under three different illumination conditions:
a) Natural illumination (NA): windows are open and sunlight enters the room. Obviously this illumination is not constant along days (due to weather conditions) and it also varies regarding the different hours of the day.
b) Infrared Illumination (IR): printed circuit board around the webcam is turned on and the remaining sources of light are disconnected. A graphic user interface has been developed in order to properly set the IRED's intensity level and to set the image involved parameters (exposure, gamma and brightness). Additionally, it is also possible to manually fully optimize them.
c) Artificial Illumination (AR): The equipment used for illumination is the following: A set of 9 cool white fluorescents uniformly distributed in order to produce the base illumination of the. A second pair of IANIRO Lilliput lights fitting 650W-3,400K tungsten halogen lamps in order to fill and smooth the well-known discontinuous fluorescent spectral emission and to provide and additional IR portion of ligth. Figure 4 shows the related portion of spectral emission in this band emitted by a set of different colour temperature halogen bulbs.

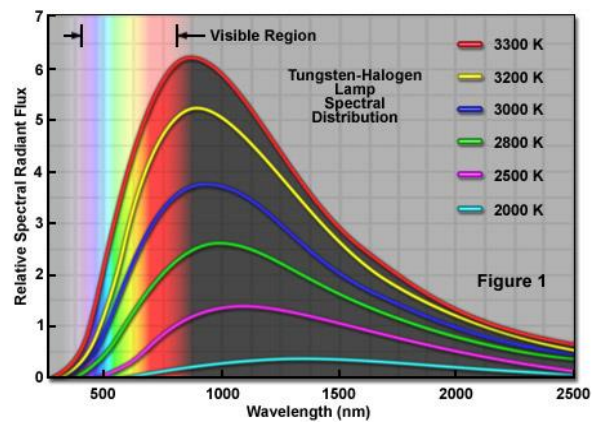

Figure 4. Relatively spectral energy distribution of different tungsten-halogen ligth sources. Zeiss courtesy.

At the beginning, high pair of power focus produced important dark shadows over the users' face. In order to solve this drawback, we had finally used a LEE 3ND 209 Filter to minimize the referred effect. This neutral density gel reduces light without affecting colour balance.

2.3 Acquisition protocol

Each user has been recorded in four different acquisition sessions performed between November of 2009 and January 2010. In this sense, distinctive changes in the haircut and/or facial hair of some subjects may be appreciated. The acquisitions have been done in the whole day from 9 AM to 5 PM, because it was getting dark after 5 PM. The average time required for the full acquisition process of a skilled user has been 10 minutes, being 15 minutes for a non skilled one. The whole set of users were acquired in two days per session.

The time slot between each session is shown in figure 5.

| Session1 (Two days) | 1 week | Session 2 (Two days) | 1 week | Session3 (Two days) | 4 weeks | Session 4 (Two days) |
|---|---|---|---|---|---|---|

Figure 5. Acquisition plan

In each illumination condition five different frontal snapshots are acquired. During the acquisition process, the user is required to look straight at the same spot. No neutral facial expression is required. Thus, different facial expressions have been collected (smiling/non smiling, open-closed and blinking eyes…etc). As glasses exhibit a fully different behavior as function of the spectrum, being transparent from the VIS to the NIR spectrum and fully opaque beyond 3μm approximately as showed in Figure 6, people wearing glasses were asked to remove them before acquisition. No other physical restriction has been taken into account in order to acquire a face image.

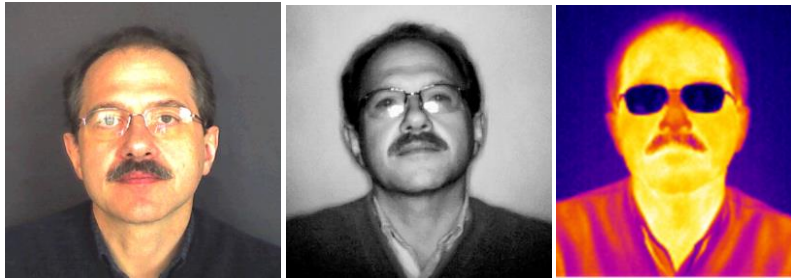

Figure 6. A set of different images in VIS, NIR and THIR of a same subject wearing glasses.

In order to reduce the correlation between consecutive acquisitions of the same session, between a couple of snapshots the user is asked to stand up, make a loop in the room, including one step which corresponds to the portion of the room close to the blackboard, and sit down again. It is worth mentioning that the thermal camera can detect a temperature increase due to this additional physical exercise.

### 2.4 Database features

Final database consists of 41 people (32 males, 9 females). Each individual contributed in four acquisition sessions (see figure 5) and provided five different snapshots in three different illumination conditions and under three image sensors. This implies a total of: 41x4x5x3x3= 7.380 images, grouped in folders shown in figure 7.

In order to normalize all the images to the same size and remove the background we have used a Viola and Jones face detector [9]. However, it was unable to segment correctly the thermal images and a new face segmentation algorithm for thermal images has been developed [10]. All the faces have been segmented and consequently resized to 100x145 pixels using bicubic interpolation.

The images in NIR spectrum are stored in lossless *.bmp files. The images from thermal camera were firstly stored to *.bmt format provided by TESTO company. This file includes VIS image, temperature matrix and metadata describing for example the outside humidity, temperature range etc. This file was processed and the image in VIS spectrum was extracted. The temperature matrix was stored to MATLAB *.mat file and also transformed to grayscale image and stored to *.bmp format.

Each file in database has an 8-letter code name. The meaning of each letter is described in Table 1.

| Letter position | 1–2 | 3–4 | 5 | 6–7 | 8 |
|---|---|---|---|---|---|
| Meaning | Personal ID | Session number | Sensor | Illumination | Sample |
| Possible values | 01–41 | S1–S4 | C – visible<br>I – near infrared<br>T – thermal | NA – natural<br>IR – infrared<br>AR – artificial | 1–5 |

Table 1. Meaning of the file code name

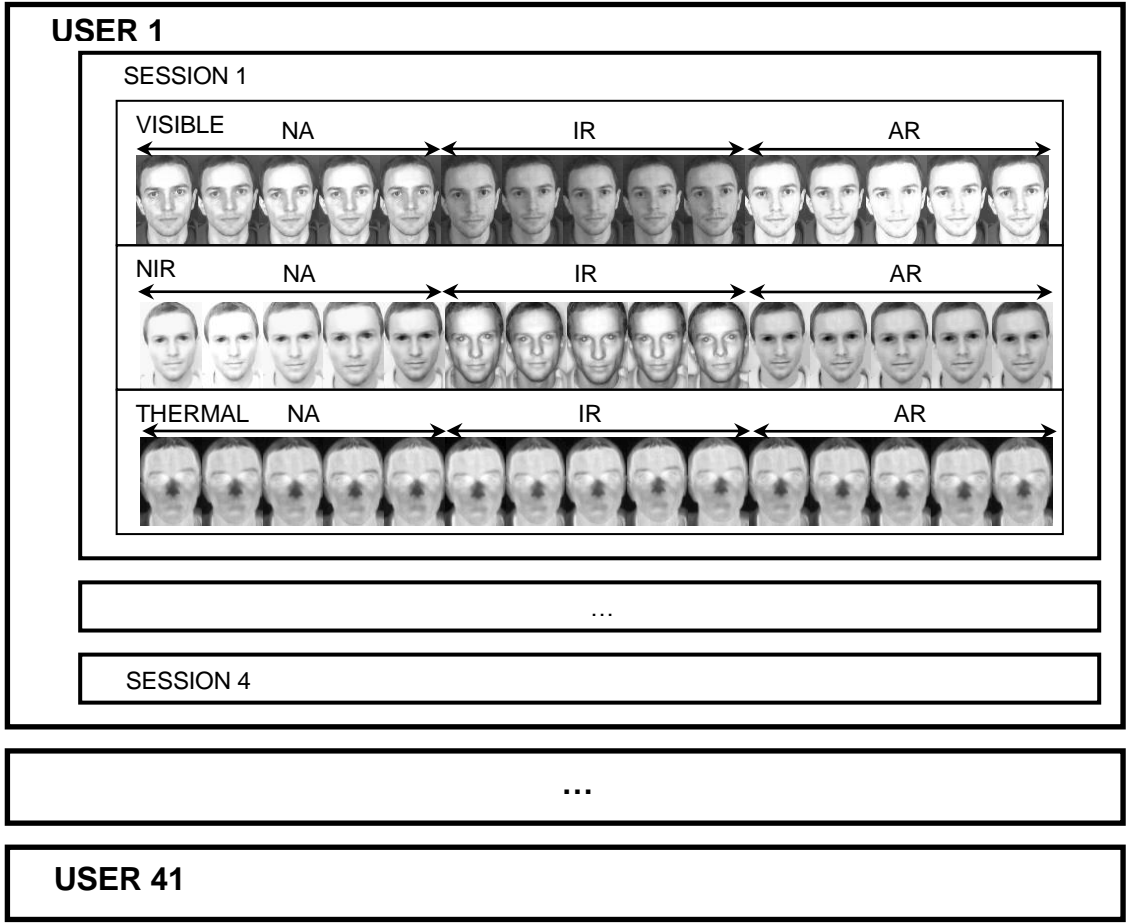

Figure 7. Database structure. For each user there are four sessions and each session contains three kinds of sensors and three different types of illumination per sensor.

### 3. EXPERIMENTAL RESULTS

In order to compare the identification rates using different sensors and illumination conditions we have used a simple feature extraction method based on Discrete Cosine Transform (DCT). According to our previous experiments [11], this method outperforms the well known eigenfaces [12] algorithm with lesser computation burden.

#### 3.1 Feature extraction algorithm

Given a face image, the first step is to perform a two dimensional DCT, which provides an image of the same size but with most of the energy compacted in the low frequency bands (upper left corner).

The Discrete Cosine Transform (DCT) is an invertible linear transform and is similar to the Discrete Fourier Transform (DFT). The original signal is converted to the frequency domain by applying the cosine function for different frequencies. After the original signal has been transformed, its DCT coefficients reflect the importance of the frequencies that are present in it. The very first coefficient refers to the signal's lowest frequency, and usually carries the majority of the relevant (the most representative) information from the original signal. The last coefficients represent the component of the signal with the higher frequencies. These coefficients generally represent greater image detail or fine image information, and are usually noisier. DCT has an advantage when compared with DFT: the coefficients are real values while DFT produces complex values. We used DCT2 (two dimensional DCT) defined by the following equations:

$$X[k,l] = \frac{2}{N} c_k c_l \sum_{m=0}^{a-1} \sum_{n=0}^{b-1} x[m,n] \cos\left[\frac{(2m+1)k\pi}{2N}\right] \cos\left[\frac{(2n+1)l\pi}{2N}\right] \quad (1)$$

where, in Equation (1) :

$$c_k, c_l = \begin{cases} \sqrt{\left(\dfrac{1}{2}\right)} & to \quad k=0, l=0 \\ 1 & to \quad k=1,2,\ldots a-1 \quad and \quad l=1,2,\ldots b-1 \end{cases} \quad (2)$$

Figure 8 summarizes the process to obtain a feature vector from a DCT transformed image.

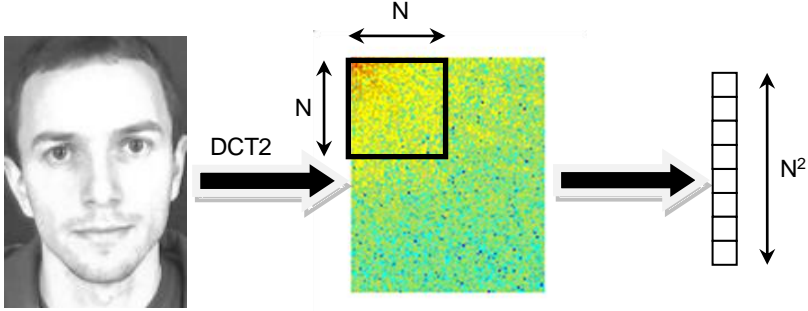

Figure 8. The process to obtain a feature vector from a face image consists of a two dimensional DCT plus a feature selection.

Feature extraction using DCT consists of selecting the coefficients around the X[0,0] coefficient (DC coefficient), where the highest discrimination capability between different people is.

We can define a zonal mask as the matrix $m(f_1, f_2) = \begin{cases} 1, & f_1, f_2 \in I_t \\ 0, & otherwise \end{cases}$, and multiply the transformed image by the zonal mask, which takes the unity value in the zone to be retained and zero on the zone to be discarded. In image coding it is usual to define the zonal mask taking into account the transformed coefficients with largest variances. In image coding the goal is to reduce the amount of bits without appreciably sacrificing the quality of the reconstructed image, and in image recognition the number of bits is not so important. The goal is to reduce the dimensionality of the vectors in order to simplify the complexity of the classifier and to improve recognition accuracy. In this paper, we will follow a frequency selection mechanism by means of discriminability criteria. The goal is to pick up those frequencies that yield a low intra-class variation and high inter-class variation. On the other hand, those frequencies that provide a high variance for inter and intra-class distributions should be discarded. The notation is the following one:

- $P$ is the number of people inside the database.
- $F$ is the number of images per person in the training subset.
- $i_{p,f}(x,y)$ is the luminance of a face image $f$ that belongs to person $p$, where $p=1,\cdots P; f=1,\cdots, F$
- $I_{p,f}(f_1, f_2) = transform\{i_{p,f}(x,y)\}$ is the DCT2 transformed image.
- $m(f_1, f_2) = \dfrac{1}{P \times F} \sum_{p=1}^{P} \sum_{f=1}^{F} I_{p,f}(f_1, f_2)$ is the average of each frequency obtained from the whole training subset images.
- $m_p(f_1, f_2) = \dfrac{1}{F} \sum_{f=1}^{F} I_{p,f}(f_1, f_2) \quad \forall p=1,\cdots P$ is the average of each frequency for each person $p$.
- $\sigma_p^2(f_1, f_2) = \dfrac{1}{F} \sum_{f=1}^{F} \left(I_{p,f}(f_1, f_2) - m_p(f_1, f_2)\right)^2$, $\forall p=1,\cdots P$ is the variance of each frequency for each person $p$.
- $\sigma^2(f_1, f_2) = \dfrac{1}{P \times F} \sum_{p=1}^{P} \sum_{f=1}^{F} \left(I_{p,f}(f_1, f_2) - m(f_1, f_2)\right)^2$, $\forall p=1,\cdots P$ is the variance of each frequency evaluated over the whole training subset.
- $\sigma_{intra}^2(f_1, f_2) = \sum_{p=1}^{P} \sigma_{p,f}^2(f_1, f_2)$ is the average of the variance of each frequency for each person.
- $\sigma_{inter}^2(f_1, f_2) = \sigma^2(f_1, f_2)$.

We will use the following measure, which is the Fisher discriminant:

$$M_1(f_1, f_2) = \frac{|m_{\text{intra}}(f_1, f_2) - m_{\text{inter}}(f_1, f_2)|}{\sqrt{\sigma^2_{\text{intra}}(f_1, f_2) + \sigma^2_{\text{inter}}(f_1, f_2)}} \qquad (3)$$

It is interesting to point out that this procedure is similar to the threshold coding used in transform image coding [13]. Nevertheless, we are using a discriminability criterion, instead of a representability criterion, which is only based on energy (the higher the frequency coefficient value, the higher its importance).
Figure 9 shows an example of the $M_1$ ratio obtained from visible images of session 1.
It is important to point out that feature selection has been done using only training samples. Thus, we have not selected frequencies using testing samples, which would provide better results, but unrealistic because feature selection must be done a priori, before classifying samples.

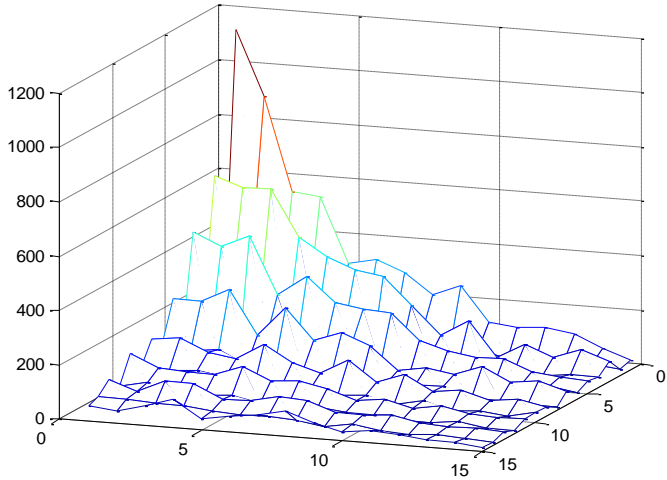

Figure 9. 15x15 first coefficients $M_1$ ratio for visible images of session 1. It is evident that the highest discrimability power is around the low frequency portion (upper left corner).

### 3.2 Classification algorithm

We have used a simple distance calculation between training and testing feature vectors of dimension $N$ using a fractional distance. We have also successfully applied this classifier in [14] for signature recognition and in [15] for speaker recognition. It is represented in equation (4):

$$d(\vec{x}, \vec{y}) = \left(\sum_{i=1}^{N}(|x_i - y_i|)^p\right)^{1/p} \qquad (4)$$

Where $i$ is the feature vector component.
For $p=2$ the equation corresponds to the Euclidean distance. When data are high dimensional, however, the euclidean distances and other Minkowsky norms (p-norm with p being an integer number, i.e, p = 1; 2; ...) seem to concentrate and, so, all the distances between pairs of data elements seem to be very similar [16] Therefore, the relevance of those distances have been questioned in the past, and fractional p-norms (Minkowski-like norms with an exponent p less than one) were introduced to fight the concentration phenomenon. In our case, we have used $p=0.5$.

We have experimentally selected the number of coefficients (vector dimension) by trial and error, selecting a window of 1x1, 2x2, 3x3, …,NxN, where the frequency coefficients have been previously ordered using the strategy defined in section 3.1.

We have used a simple method because the experiments are quite time consuming. For each feature vector dimension we have executed the algorithm, we have studied hundreds of feature vector dimensions for each condition, and this implies thousands of executions. If a more sophisticated method were used, this would imply, probably, to train a complex algorithm for each studied feature vector dimension. This would be impractical from the computational burden point of view. In fact, in its current version, we required several weeks to work out the whole experimental section.

In addition, we were looking for a method with few parameters because a more complex algorithm can require fine tuning, and this fine tuning could be different for each spectral band. Thus, in this case, it would be difficult to know if one spectral band provides better results due to different tuning or to the frequency itself. Our suggested method is so simple and effective that we did not require any fine tuning.

### 3.3 Experimental results with different illumination conditions

In this section we compare the identification rates for the visible (VIS), Near infrared (NIR) and thermal (TH) sensors for natural (NA, figure 10), artificial (AR, figure 11) and infrared (IR, figure 12) light. These experimental results have been obtained by training with session 1 & 2 and testing with session 3 as function of N (see figure 8). Thus, the number of selected coefficients for each point in these plots is $N^2$.

These figures reveal several general interesting facts:
- Feature selection is indeed important, because a number of coefficients that is too large decreases the identification rate.
- Different sensors provide a different number of optimal feature dimension $N^2$.

Figure 10 shows that:
- The NIR sensor provides lower identification rates than visible and thermal ones, which provide similar rates. In addition, the optimal feature vector size is more critical, because identification rates drop quickly when moving away from the optimal point.
- The TH sensor requires a lesser amount of coefficients to reach the highest identification rate, and the identification rate drops slower than for visible sensor.

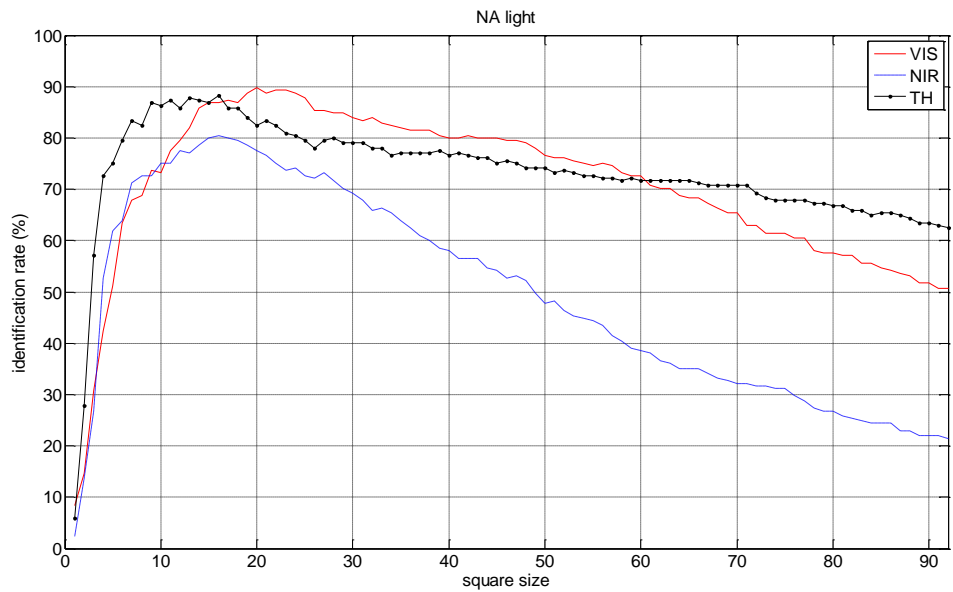

Figure 10. Identification rate as function of the square size (N) of selected coefficients for visible (VIS), near infrared (NIR) and thermal (TH) sensors for natural (NA) illumination.

Figure 11 shows that:
- All the sensors provide nearly similar results, although visible sensor outperforms the other ones.
- Optimal feature vector size selection is less critical for the VIS sensor than for the other ones because a large range of $N^2$ values produce the highest achievable identification rate.

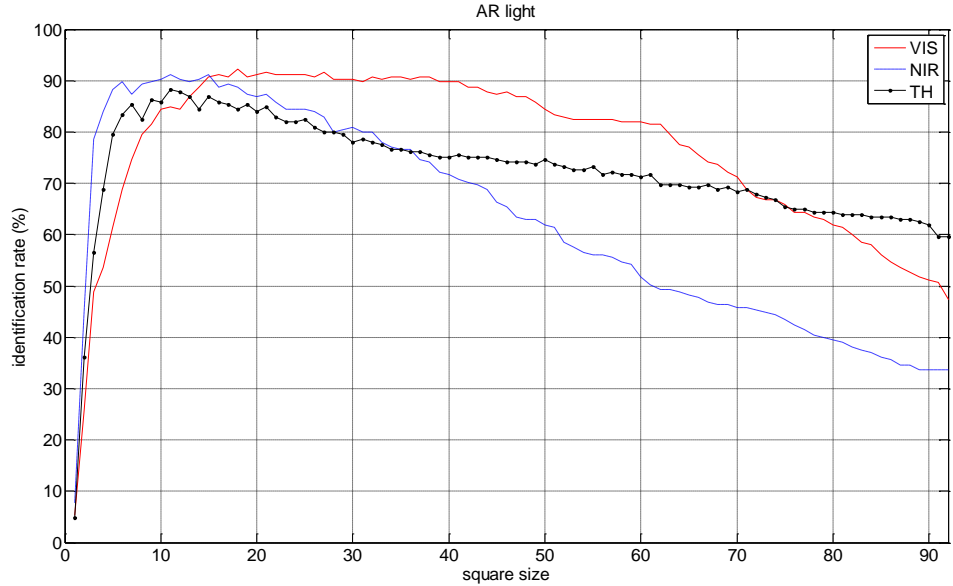

Figure 11. Identification rate as function of the square size (N) of selected coefficients for visible (VIS), near infrared (NIR) and thermal (TH) sensors for artificial (AR) illumination.

Figure 12 shows that:
- NIR sensor provides the best behavior and the VIS sensor fails to provide a reasonable identification rate. This makes sense considering that infrared illumination in the proposed scenario for a visible sensor is equivalent to a near-dark scene.
- TH and NIR provide similar behavior, although TH sensor results drop faster beyond the optimal value.

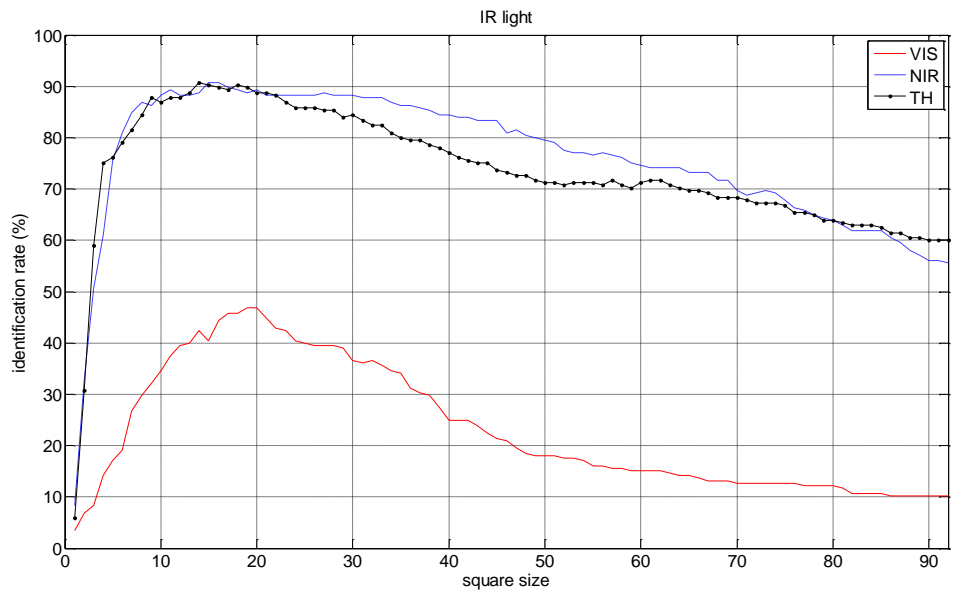

Figure 12. Identification rate as function of the square size (N) of selected coefficients for visible, near infrared and thermal sensors for near infrared (IR) illumination.

   3.4  Experimental results for a specific sensor
   In this section we compare the identification rates for a specific sensor regarding of the different illumination conditions. We have studied the VIS sensor (figure 13), the NIR (figure 14) and the TH (figure 15) for natural (NA), artificial (AR) and infrared (IR) illumination.
   Figure 13 reveals that:
- VIS sensor performs better with artificial illumination. This makes sense because the variation along acquisition sessions is smaller than when using natural light, which varies from day to day.

- Optimal feature selection value is more critical when using natural light when compared to artificial light.
- VIS sensor fails when using NIR illumination. This is due to the acquisition conditions for this scenario, which is almost dark for a visible sensor.

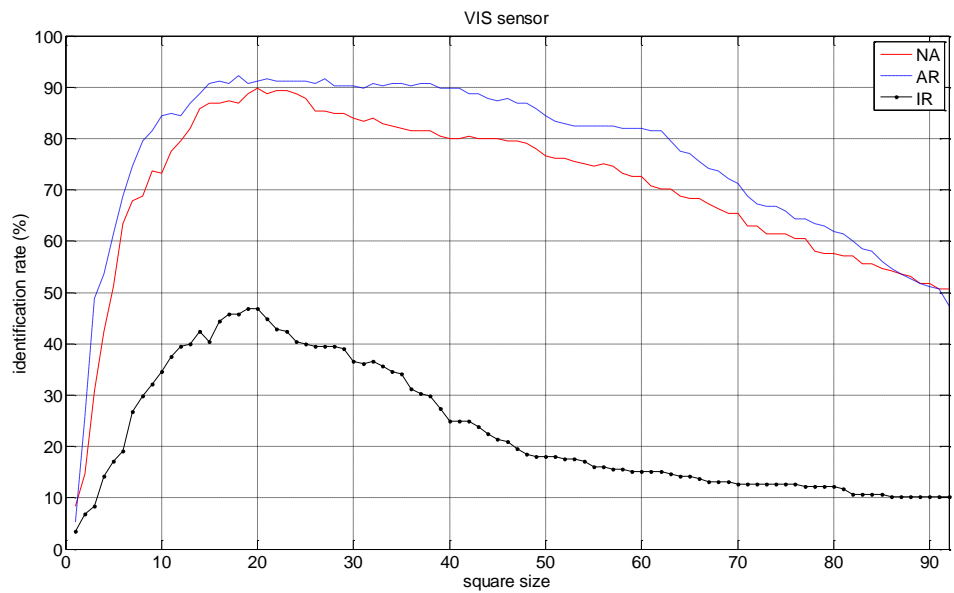

Figure 13. Identification rate as function of the square size (N) of selected coefficients for Visible sensor and natural (NA), artificial (AR) and near infrared (NIR) illumination.

Figure 14 reveals that:
- IR sensor performs similarly well with AR and IR illumination, and around 10% worse when evaluated with natural light. This can be due to the larger variability when analyzing faces with natural light.
- Feature selection is less critical when using IR illumination. This is reasonable considering that NIR sensors should perform optimally with IR illumination.

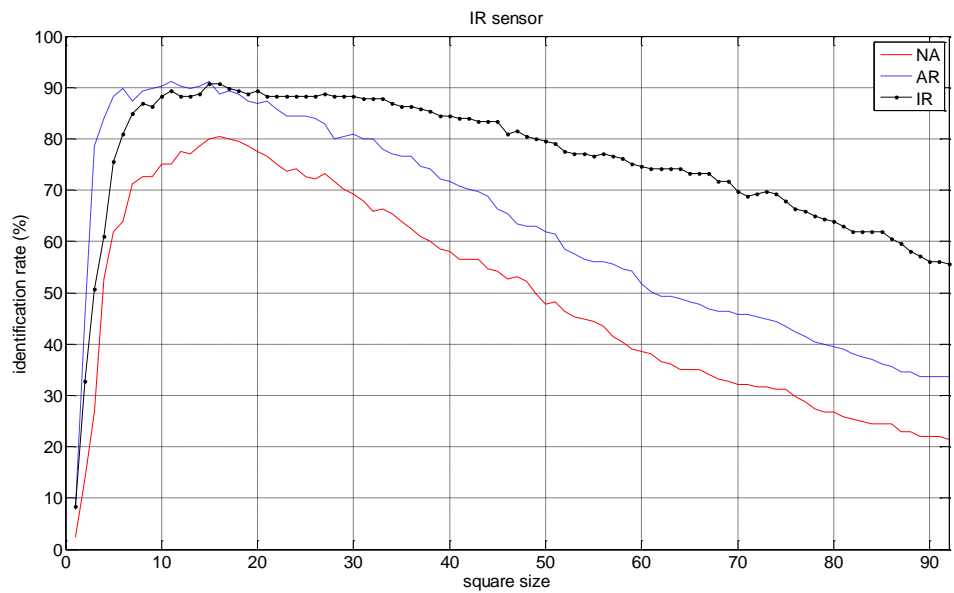

Figure 14. Identification rate as function of the square size (N) of selected coefficients for Infrared sensor and natural (NA), artificial (AR) and near infrared (NIR) illumination.

Figure 15 shows an expected conclusion:

- TH sensor performs almost the same with all the illuminations studied. This is reasonable considering that thermal cameras do not measure the light reflection on the face. They measure the heat emission of the body. In fact, they could perfectly work in fully darkness because the illumination is irrelevant.

It is important to point out that although there are small variations between the three illumination conditions, they are not due to illumination. The motivation is the inherent variability of the acquired subject from day to day and acquisition to acquisition. If the subject were an inanimate object with a fix temperature along the different acquisitions, the behaviour shown in figure 12 would be the same under the three illuminations. However, a human being cannot fulfil this property.

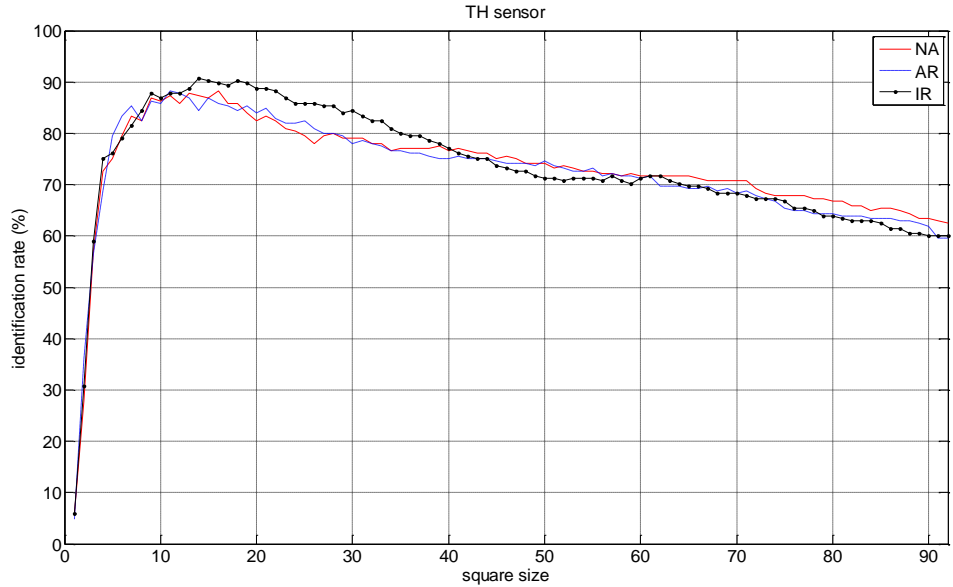

Figure 15. Identification rate as function of the square size (N) of selected coefficients for Thermal sensor and natural (NA), artificial (AR) and near infrared (NIR) illumination.

Table 2 summarizes the optimal results and the optimal feature vector dimension (when evaluated from 1x1, 2x2,… NxN) for different sensors and illumination conditions. This table reveals similar identification rates for all the sensors, although the thermal one requires a lower number of coefficients. In addition the visible sensor provides low identification rates when using IR illumination for the reasons previously commented.

|        | Illumination | | | | | |
|--------|--------------|--|--|--|--|--|
|        | NA | | IR | | AR | |
| Sensor | Identification | Coefficients | Identification | Coefficients | Identification | Coefficients |
| VIS | 89.76 | 20x20 | 46.83 | 19x19 | 92.20 | 18x18 |
| NIR | 80.49 | 16x16 | 90.73 | 15x15 | 91.22 | 11x11 |
| TH  | 88.29 | 16x16 | 90.73 | 14x14 | 88.29 | 11x11 |

Table 2. Optimal results for visible (VIS), near infrared (NIR) and thermal (TH) sensor under natural (NA), infrared (IR) and artificial (AR) illumination conditions. Experimental conditions are the same of previous figures 10 to 15. The selected number of coefficients is also represented.

### 3.5 Experimental results in mismatch conditions

Using the setup of previous sections we have studied the identification rates regarding the different illumination conditions for training and testing. Table 3 shows the experimental results when using 20x20 coefficients for VIS, NIR and TH respectively. The models have been computed using session 1 and 2 and the testing has been done with session 3 and 4 separately.

Although it is possible to trade-off an optimal feature vector dimension for each scenario we decided to select a fix window size of 20x20 coefficients. According to previous plots (figures 10 to 15) this tends to benefit the identification rates of the VIS sensor. Nevertheless, the goal of this table is to study the mismatch illumination effect between training and testing conditions, rather than to find the highest identification rate for each scenario.

Due to the bad results obtained specially when using IR illumination we have decided to use some normalization procedure. The image has been normalized prior to DCT2. The normalization maps the values in intensity image into new values such that 1% of data is saturated at low and high intensities of the image. This increases the contrast of the normalized image. Thus, table 3 includes experimental results with and without normalization.

Face recognition in new spectral bands imply new problems that must be addressed. Nevertheless, the solution should be specific for each spectral band. For instance, while general temperature can rise in thermal images, the relative difference between different portions of the face can remain similar, because the hottest point will always be related to the vein positions, and this remains the same. On the other hand, we trained with sessions 1 and 2 and tested with sessions 3 and 4. Thus, the experimental results are affected by the time evolution. Nevertheless we have applied a feature selection algorithm that looks for low intra-class variation and high inter-class variation. Thus, stability along time is achieved by means of feature selection (see section 3.1), which is different for each spectral band.

|        |               |        | Test |      |      |      |      |      |
|--------|---------------|--------|------|------|------|------|------|------|
|        |               |        | NA   |      | IR   |      | AR   |      |
| Sensor | Normalization | Train  | 3    | 4    | 3    | 4    | 3    | 4    |
| VIS    | NO            | NA 1&2 | 89.8 | 84.4 | 51.2 | 60   | 85.9 | 82.9 |
| VIS    | YES           | NA 1&2 | 90.2 | 83.9 | 80.5 | 79   | 86.8 | 83.9 |
| VIS    | NO            | IR 1&2 | 85.9 | 82.4 | 46.8 | 61   | 92.2 | 85.9 |
| VIS    | YES           | IR 1&2 | 86.8 | 83.9 | 90.2 | 91.7 | 90.7 | 86.3 |
| VIS    | NO            | AR 1&2 | 90.2 | 82   | 57.6 | 60   | 91.2 | 87.3 |
| VIS    | YES           | AR 1&2 | 89.8 | 81.5 | 85.4 | 84.9 | 91.7 | 89.8 |
| NIR    | NO            | NA 1&2 | 77.6 | 89.8 | 19   | 21   | 79   | 85.4 |
| NIR    | YES           | NA 1&2 | 82   | 93.7 | 44.4 | 38   | 85.4 | 87.8 |
| NIR    | NO            | IR 1&2 | 70.7 | 54.1 | 89.3 | 89.3 | 63   | 61.5 |
| NIR    | YES           | IR 1&2 | 74.1 | 57.1 | 94.1 | 95.1 | 62   | 54.6 |
| NIR    | NO            | AR 1&2 | 78.5 | 81   | 25   | 26.3 | 86.8 | 88.8 |
| NIR    | YES           | AR 1&2 | 85.4 | 86.3 | 49.3 | 49.3 | 89.8 | 88.3 |
| TH     | NO            | NA 1&2 | 82.4 | 80.5 | 82   | 78   | 83.4 | 77.1 |
| TH     | YES           | NA 1&2 | 82.4 | 80   | 82.9 | 79.5 | 84.4 | 78.5 |
| TH     | NO            | IR 1&2 | 82.4 | 78   | 88.8 | 78   | 84.9 | 75.1 |
| TH     | YES           | IR 1&2 | 82.4 | 76.6 | 88.3 | 78.5 | 84.9 | 75.1 |
| TH     | NO            | AR 1&2 | 81.5 | 76.1 | 82.4 | 80   | 83.9 | 81   |
| TH     | NO            | AR 1&2 | 82   | 78.5 | 82   | 79.5 | 84.9 | 81   |

Table 3. Identification rates (%) under different illuminations, sensors and normalization conditions

Table 3 reveals the following aspects:
- IR sensor provides the best result, which is 94.1% identification rate. This experimental result is in agreement with the conclusion of our previous paper [3], because NIR images have higher entropy than the other ones.
- Looking at the standard deviation (std) and mean value (m) of the experimental results of table 3 for a specific sensor we obtain: m=81.6 and std= 12.4 for visible sensor, m=69.6 and std=22.8 for near infrared sensor and m=80.9 and std=3.2 for Thermal sensor. Thus, thermal image recognition rates are more stable than the other sensors.
- Image normalization is important for the case of illumination mismatch when using the visible and near infrared sensor, and less important for the thermal one.

### 3.6 Experimental results using multi-sensor score fusion

In this section we combine the scores provided by different sensors in order to improve recognition accuracies. The existing fusion levels [17] are sensor, feature, score and decision. Some papers use image fusion and then they perform the recognition over this fused image [18-22]. This is known as "sensor fusion". Another possibility is decision fusion [23]. In our paper, we will use a score combination. Some papers have also studied this possibility [24-27]. In fact [25] studies the sensor, feature and score level and founds that data fusion at score level outperforms the other ones when combining visible and thermal images. However [26] studies the fusion of visible and near infrared images and founds slightly better accuracies when fusing images than applying other fusion levels. To the best of our knowledge there is no paper devoted to visible, near infrared and thermal images simultaneously. The fusion scheme is presented in figure 16.

This kind of fusion is also known as confidence or opinion level. It consists of the combination of the scores provided by each matcher. The matcher just provides a distance measure or a similarity measure between the input features and the models stored on the database.

Before opinion fusion, normalization must be done when the scores provided by different classifiers do not lie in the same range. In our case, we experimentally found that this normalization is not necessary because the three classifiers studied gave similar range.

After the normalization procedure, several combination schemes can be applied [17]. The combination strategies can be classified into three main groups:
  a) Fixed rules: All the classifiers have the same relevance. An example is the sum of the outputs of the classifiers. That is: let $o_1$ and $o_2$ be the outputs of classifiers number 1 and 2 respectively. For example, a fixed combination rule yields the combined output $O = (o_1 + o_2)/2$
  b) Trained rules: Some classifiers should have more relevance on the final result. This is achieved by means of some weighting factors that are computed using a training sequence. For instance: $O = \omega_1 o_1 + \omega_2 o_2 = \omega_1 o_1 + (1 - \omega_1) o_2$
  c) Adaptive rules: The relevance of each classifier depends on the instant time. This is interesting for variable environments. That is: $O = \omega_1(t) o_1 + (1 - \omega_1(t)) o_2$ .For instance, a system that detects a low illumination scene and then weighs more the thermal score.

$$O_j = \sum_{i=1}^{N} \omega_i o_{ij}$$

The most popular combination scheme is the weighted sum:
Where the weights can be fixed, trained or adaptive.

In this paper we will use a fixed rule scheme as well as a trained rule, although in our case the purpose of the trained rule is to evaluate the weights assigned to each classifier, rather than to maximize the identification rate. In fact, trained rules should be done with a development set different than the test set. Otherwise the experimental results are unrealistic.

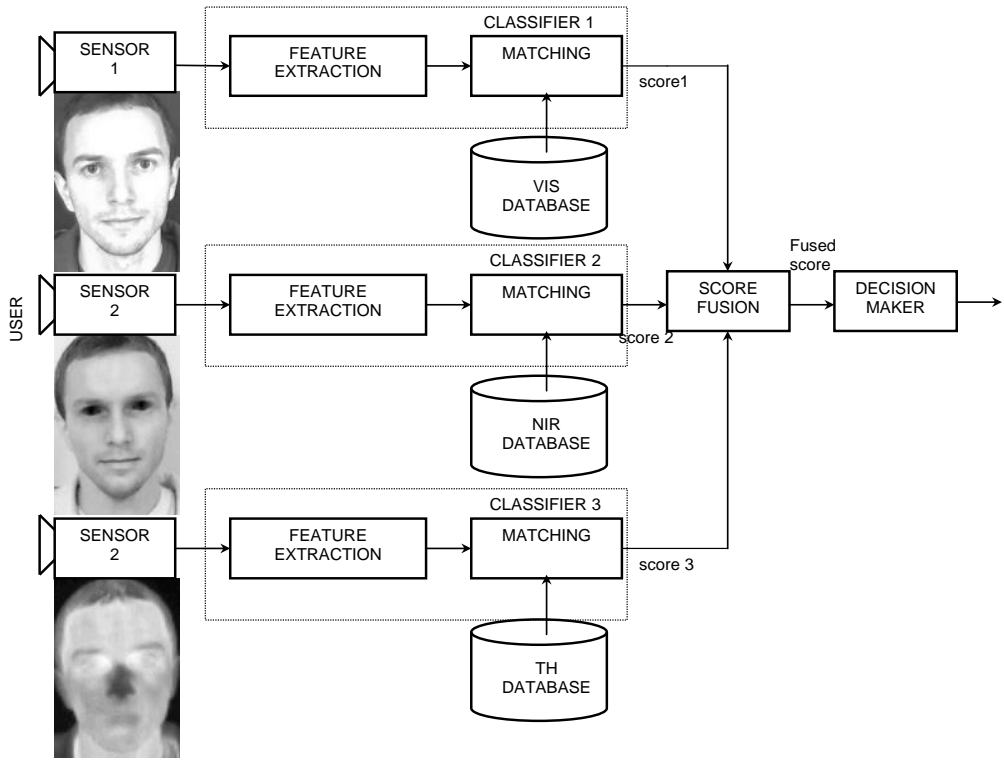

Figure 16. Multi-sensor fusion scheme at score level.

Table 4 shows the identification rates under different training and testing conditions for a fixed rule using the same weight for all the classifiers.

|  |  |  | Test |  |  |  |  |  |
|---|---|---|---|---|---|---|---|---|
|  |  |  | NA | | IR | | AR | |
| Sensors | Normalization | Train | 3 | 4 | 3 | 4 | 3 | 4 |
| VIS&NIR | NO | NA 1&2 | 91.7 | 96.1 | 51.2 | 58 | 91.7 | 92.7 |
| VIS&NIR | YES | NA 1&2 | 94.1 | 96.6 | 82 | 81.5 | 94.6 | 94.6 |
| VIS&NIR | NO | IR 1&2 | 93.7 | 90.7 | 91.7 | 92.7 | 91.7 | 91.7 |
| VIS&NIR | YES | IR 1&2 | 93.7 | 89.8 | 97.6 | 98 | 96.1 | 91.2 |
| VIS&NIR | NO | AR 1&2 | 92.7 | 91.7 | 49.3 | 52.2 | 95.1 | 94.6 |
| VIS&NIR | YES | AR 1&2 | 96.1 | 93.7 | 86.8 | 82.4 | 97.07 | 96.1 |
| VIS&TH | NO | NA 1&2 | 91.7 | 87.3 | 85.9 | 82.9 | 95.1 | 88.8 |
| VIS&TH | YES | NA 1&2 | 94.1 | 88.3 | 91.7 | 87.8 | 95.1 | 88.3 |
| VIS&TH | NO | IR 1&2 | 96.1 | 89.3 | 90.2 | 84.4 | 96.6 | 90.2 |
| VIS&TH | YES | IR 1&2 | 96.6 | 86.8 | 97.1 | 94.1 | 98 | 90.7 |
| VIS&TH | NO | AR 1&2 | 95.6 | 89.3 | 83.4 | 82.9 | 98 | 93.7 |
| VIS&TH | YES | AR 1&2 | 95.6 | 89.3 | 93.2 | 91.2 | 98 | 93.7 |
| NIR&TH | NO | NA 1&2 | 91.7 | 96.1 | 85.9 | 74.1 | 95.1 | 96.6 |
| NIR&TH | YES | NA 1&2 | 94.1 | 97.6 | 91.7 | 82 | 97.1 | 97.1 |
| NIR&TH | NO | IR 1&2 | 96.6 | 89.3 | 96.1 | 96.6 | 92.2 | 87.8 |
| NIR&TH | YES | IR 1&2 | 94.6 | 84.4 | 98.5 | 99 | 90.2 | 85.4 |
| NIR&TH | NO | AR 1&2 | 93.7 | 94.6 | 77.1 | 67.8 | 97.1 | 93.7 |
| NIR&TH | YES | AR 1&2 | 96.1 | 96.1 | 84.4 | 80.5 | 98.5 | 93.7 |
| VIS&NIR&TH | NO | NA 1&2 | 94.6 | 97.1 | 82.9 | 76.6 | 97.1 | 96.1 |
| VIS&NIR&TH | YES | NA 1&2 | 94.6 | 97.6 | 94.1 | 91.7 | 97.6 | 97.1 |
| VIS&NIR&TH | NO | IR 1&2 | 98.5 | 95.1 | 98 | 97.6 | 98.5 | 97.1 |
| VIS&NIR&TH | YES | IR 1&2 | 98 | 94.6 | 98.5 | 100 | 99.5 | 96.6 |
| VIS&NIR&TH | NO | AR 1&2 | 97.1 | 95.1 | 80.5 | 74.6 | 99.5 | 97.1 |
| VIS&NIR&TH | YES | AR 1&2 | 98.5 | 98.5 | 93.7 | 91.7 | 99.5 | 98 |

Table 4. Identification rate for the combination of two and three sensors under different illumination conditions (NA=Natural, IR=Infrared, AR=Artificial).

When combining two classifiers using a trained rule, a trial and error procedure must be done to set up the optimal value of the weighting factor. Figure 17 shows the identification rates as function of the weighting factor *alpha*, where the combination function is:

$$d = alpha \times d_{VIS} + (1 - alpha) \times d_{NIR}$$

It is interesting to point out that for *alpha*=1 the combination consists of the visible classifier distance alone $d_{VIS}$, while *alpha*=0 fully removes the effect of the visible classifier, the classification being based on near infrared sensor distance alone $d_{NIR}$. Thus, for *alpha*=0 we obtain 89.8% identification rate and for *alpha*=1 84.4%. In the middle, there is an area that provides higher recognition rates (up to 95.6%) due to the combination of distances.

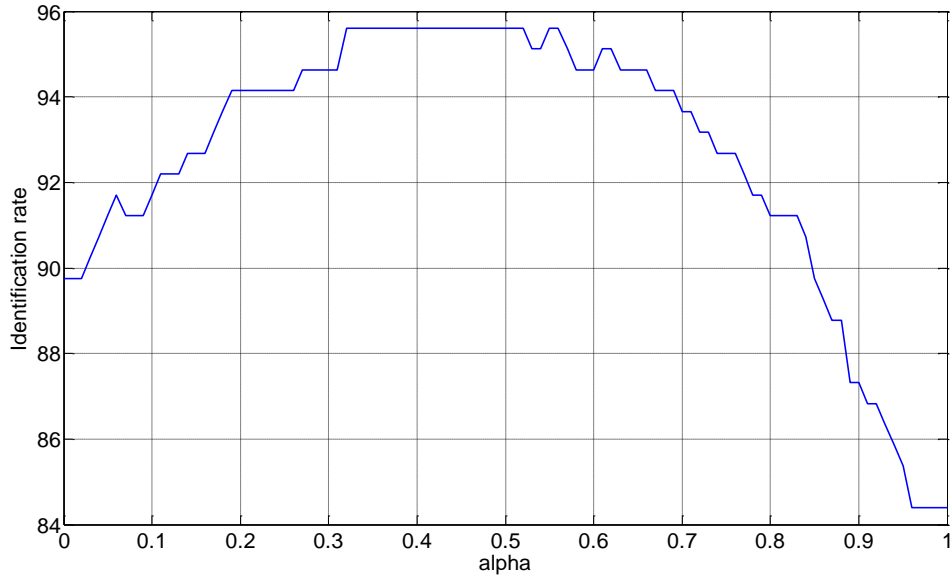

Figure 17. Trained rule combining VIS and NIR classifiers for NA illumination for training and testing, session 4.

When combining three classifiers we can generalize the previous procedure using the following combination function:
$$d = alpha \times d_{VIS} + beta \times d_{NIR} + (1 - alpha - beta) \times d_{TH}$$
In this case we should trade-off two parameters and the graphical representation is a three dimensional plot, such as the one shown in figure 18. This three dimensional plot is not very informative due to the limitations of three dimensional representations and an alternative is to represent its contour plot. A contour plot are the level curves of the bidimensional matrix formed by giving values to the two parameters *alpha* and *beta*. For the sake of simplicity only a few level curves are plot, as well as a black dot that indicates the highest value. Some interesting remarks about this kind of plot are the following:

- In fact, the addition of the three weighting factors should be one. However, in order to avoid discontinuities and sudden gradients, we have filled up a whole matrix with $alpha, beta \in [0,1]$ using increments of 0.01. Thus, 100 values have been worked out for each variable.
- *Alpha*=100 implies *beta*=0. Thus, the combined system consists of the visible sensor alone.
- *Beta*=100 implies alpha=0. Thus, the combined system consists of the near infrared sensor alone.
- *Alpha=beta*=0 implies that the combined system consists of the thermal sensor alone.
- *Alpha=beta*=33 implies that the three systems are equally weighted in the averaged distance computation.
- *Alpha* and *beta* adjustments on the diagonal line depicted in each of the figures 19 and 20 imply that the thermal sensor is not used. The closest the optimal point is to this line, the lesser the weight of the thermal system. Adjustment points far from this diagonal imply a strong weight on the thermal system.

Observing the 18 plots of figures 19 and 20 it is clearly seen that the three systems are almost equally important in the weighting process. There is only one exception, which is the second plot of figure 19. In this case *alpha*=33, *beta*=0. Thus, near infrared images are ignored and thermal images are weighted two times more than visible ones. This is reasonable considering the identification rates of each sensor alone (see table 3: VIS=60%, NIR=21%, TH=78%). Using these optimal combination values the identification rate reaches 84.9%.

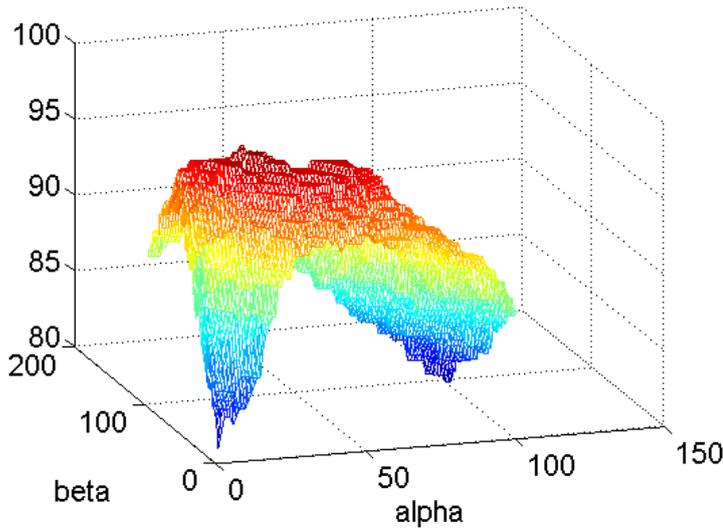

Figure 18. Example of trained rule identification rates combining three classifiers.

Figure 19 shows the contour plots as well as the maximum identification rate for the VIS, NIR and TH combination from top down and left to right for the following training and testing illumination conditions: NA-NA, NA-IR, NA-AR, IR-NA, IR-IR, IR-AR and AR-NA, AR-IR, AR-AR for session 4 and unnormalized feature vectors. Figure 20 represents the experimental results under the same illumination conditions for the normalized feature vectors case.

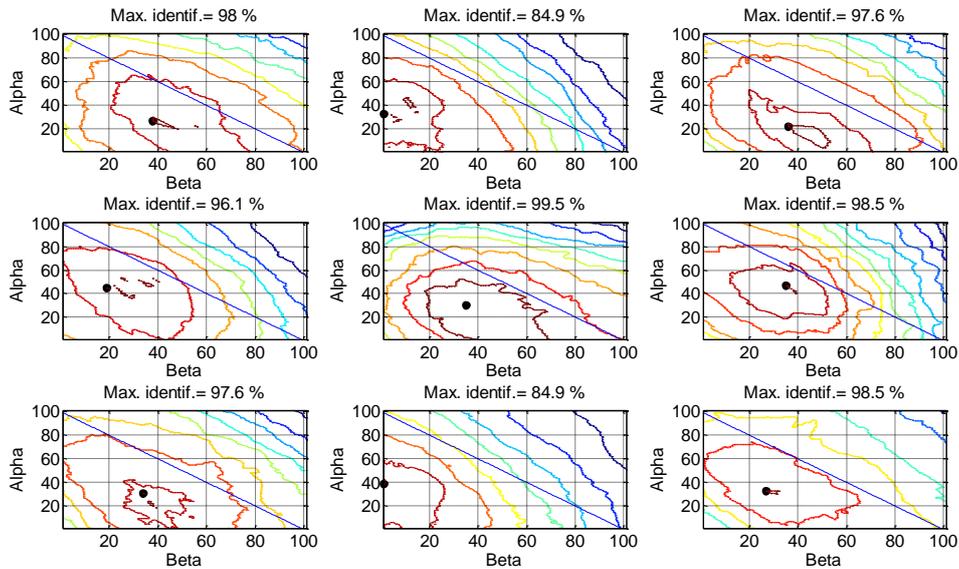

Figure 19. Contour plots when combining VIS, NIR and TH sensors under the following training and testing illumination conditions: NA-NA, NA-IR, NA-AR, IR-NA, IR-IR, IR-AR and AR-NA, AR-IR, AR-AR for session 4 and unnormalized feature vectors.

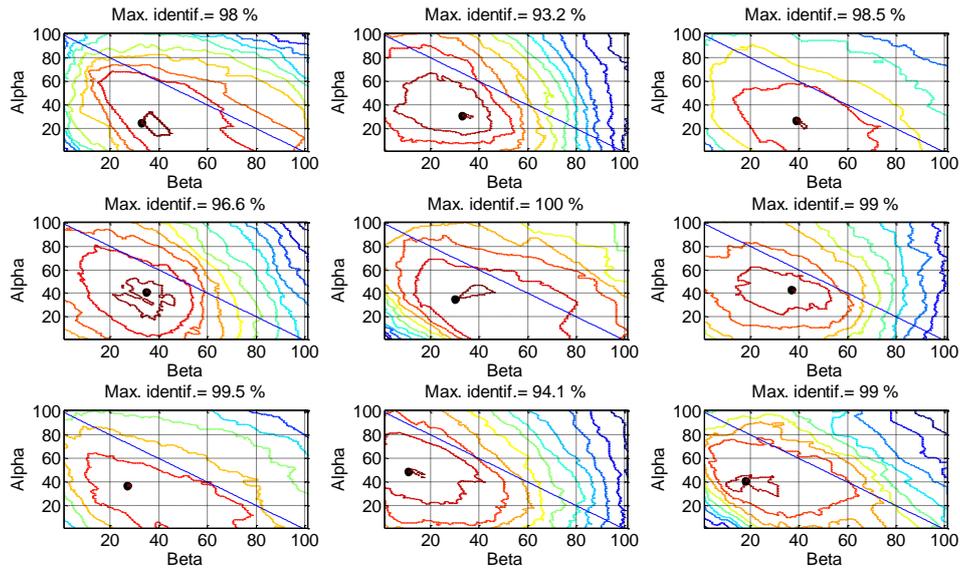

Figure 20. Contour plots when combining VIS, NIR and TH sensors under the following training and testing illumination conditions: NA-NA, NA-IR, NA-AR, IR-NA, IR-IR, IR-AR and AR-NA, AR-IR, AR-AR for session 4 and normalized feature vectors.

## 4. CONCLUSIONS

In this paper a new face database has been presented. To the best of our knowledge, this is the first database that consists of visible, near infrared and thermal images acquired simultaneously and under different illumination conditions (natural, near infrared and artificial).

The main conclusions about face recognition using a single sensor are:
- The three sensors studied can provide good identification rates.
- The highest identification rate has been obtained for NIR sensor under NIR illumination conditions.
- Thermal sensor is more stable along different illumination mismatch, as expected, and it also provides good enough identification rates, and requires smaller number of coefficients. In addition, optimal feature selection is less critical than for the other sensors.
- In average, visible sensor provides higher identification rates.

The main conclusions when fusing two or three sensors are:
- The combination improves the identification rates. The best system alone provides 95.1% identification rate, and the combined system reaches 100% in a particular scenario.
- In general the three sensors are almost equally important, because a quite balanced weighting factor is obtained by exhaustive trial and error of the whole set of weighting combinations.
- Normalized feature vectors always outperform the un-normalized system for the trained combination rule, and it is slightly worse in 3 of 18 cases for the fixed combination rule.
- When studying the three sensors simultaneously we have not found any couple of redundant sensors. The combined system takes advantage of the three spectral bands. In addition, the combined system is more robust in front of illumination mismatch.

## 5. ACKNOWLEDGEMENTS

This work has been supported by FEDER, MEC, TEC2009-14123-C04-04, KONTAKT-ME 10123, SIX (CZ.1.05/2.1.00/03.0072), CZ.1.07/2.3.00/20.0094 and VG20102014033.